\documentclass[conference]{IEEEtran}
\IEEEoverridecommandlockouts
\usepackage{cite}
\usepackage{amsmath,amssymb,amsfonts}
\usepackage{algorithmic}
\usepackage{graphicx}
\usepackage{textcomp}
\usepackage{soul}
\usepackage{color, xcolor}
\usepackage[linesnumbered,ruled,vlined]{algorithm2e}
\usepackage{geometry} 
\def\BibTeX{{\rm B\kern-.05em{\sc i\kern-.025em b}\kern-.08em
    T\kern-.1667em\lower.7ex\hbox{E}\kern-.125emX}}
\begin{document}

\title{AG-MPBS: a Mobility-Aware Prediction and Behavior-Based Scheduling Framework for Air-Ground Unmanned Systems}

\author{
    \IEEEauthorblockN{\textbf{Tianhao Shao}}
    \IEEEauthorblockA{\textit{School of Software} \\
    Northwestern Polytechnical University \\
    Xi'an, China \\
    tianhaoshao@mail.nwpu.edu.cn}
    \and
    \IEEEauthorblockN{\textbf{Kaixing Zhao}$^{*}$}\thanks{* Corresponding author.}
    \IEEEauthorblockA{\textit{School of Software} \\
    Northwestern Polytechnical University \\
    Xi'an, China \\
    kaixing.zhao@nwpu.edu.cn}
    \and
    \IEEEauthorblockN{\textbf{Feng Liu}}
    \IEEEauthorblockA{\textit{School of Software} \\
    Northwestern Polytechnical University \\
    Xi'an, China \\
   2229109248@mail.nwpu.edu.cn}
   \and
    \IEEEauthorblockN{\textbf{Lixin Yang}}
    \IEEEauthorblockA{\textit{Energy Development Research Institute} \\
    China Southern Power Grid \\
    Guangzhou, China \\
  1031thj@163.com}
   \and
    \IEEEauthorblockN{\textbf{Bin Guo}}
    \IEEEauthorblockA{\textit{School of Computer Science} \\
    Northwestern Polytechnical University \\
    Xi'an, China \\
   guob@nwpu.edu.cn}
}

\maketitle

\begin{abstract}
As unmanned systems such as Unmanned Aerial Vehicles (UAVs) and Unmanned Ground Vehicles (UGVs) become increasingly important to applications like urban sensing and emergency response, efficiently recruiting these autonomous devices to perform time-sensitive tasks has become a critical challenge. This paper presents MPBS (Mobility-aware Prediction and Behavior-based Scheduling), a scalable task recruitment framework that treats each device as a recruitable “user”. MPBS integrates three key modules: a behavior-aware KNN classifier, a time-varying Markov prediction model for forecasting device mobility, and a dynamic priority scheduling mechanism that considers task urgency and base station performance. By combining behavioral classification with spatiotemporal prediction, MPBS adaptively assigns tasks to the most suitable devices in real time. Experimental evaluations on the real-world GeoLife dataset show that MPBS significantly improves task completion efficiency and resource utilization. The proposed framework offers a predictive, behavior-aware solution for intelligent and collaborative scheduling in unmanned systems.
\end{abstract}

\begin{IEEEkeywords}
Unmanned Systems, Agent Scheduling, Mobility Prediction, Time-varying Markov Model, Priority-based Scheduling
\end{IEEEkeywords}

\section{Introduction}

With the rapid advancement of autonomous technologies, unmanned systems such as Unmanned Aerial Vehicles (UAVs) and Unmanned Ground Vehicles (UGVs) have become essential tools in various domains, including urban sensing~\cite{xu2024path}, environmental monitoring~\cite{khosiawan2016system}, disaster response~\cite{ajith2023intelligent}, and logistics~\cite{zhang2024learned, rinaldi2024comprehensive}. As their deployment scales up, one of the emerging challenges is how to dynamically and intelligently recruit these unmanned devices to perform distributed, time-sensitive tasks in an efficient and resource-aware manner~\cite{xu2024deadline, li2020multiobjective, gao2022uav}.

Traditional task assignment and resource scheduling frameworks are primarily designed for human-involved mobile crowdsensing (MCS)~\cite{ganti2011mobile}, where users voluntarily accept tasks. These approaches fall short when applied to unmanned systems, which differ significantly in mobility patterns~\cite{song2010limits}, availability models~\cite{zheng2008understanding}, and response behaviors~\cite{yuan2011driving}. Moreover, classical scheduling algorithms such as First-In-First-Out (FIFO), Earliest Deadline First (EDF)~\cite{xu2024deadline}, and Least Slack Time First (LSF)~\cite{park2014effective} lack the capability to adapt to dynamic environments, as they do not consider the real-time spatiotemporal availability of autonomous devices~\cite{liu2024real, khosiawan2019task}.

To address these limitations, this paper introduces a novel paradigm of \textit{unmanned device recruitment}, where each UAV or UGV is conceptualized as a "user" to be recruited based on its predicted future position, behavioral characteristics, and operational status~\cite{xu2024path, nannini2006analysis}. The key idea is to treat device coordination as a recruitment problem and solve it through spatiotemporal modeling and intelligent scheduling~\cite{weinstein2007uav, li2023heuristic}.

Specifically, we propose a recruitment model that integrates K-Nearest Neighbor (KNN) classification~\cite{li2010mining} to capture the behavior types of devices, such as periodic mobility or random movement, and a Time-Varying Markov Prediction (TVMP) model~\cite{ashbrook2003using, marin2015context} to predict future device locations in different time windows. This enables the system to proactively identify which devices are likely to be available near the task location and when. In addition, we develop a base-station-centric scheduling framework to orchestrate task broadcasting, collect recruitment feedback, and make decisions based on priority. Each base station maintains recruitment factors and performance metrics, such as successful task completions and response speed~\cite{liu2021theory, duan2020dynamic, chen2024delivery}, to guide future task assignments. A performance-aware scheduling strategy is then applied to maximize task success under constrained resources~\cite{li2020multiobjective, yu2019game}.

In summary, the main contributions of this paper are as follows:
\begin{itemize}
  \item[(1)] Design of MPBS, a mobility-aware prediction and behavior-based scheduling framework for heterogeneous unmanned systems (e.g., UAVs and UGVs) to support dynamic collaborative task execution.
  \item[(2)] Development of a behavior-aware agent recruitment algorithm that integrates K-Nearest Neighbor (KNN) classification and time-varying Markov-based mobility prediction.
  \item[(3)] Comprehensive experimental evaluation on the real-world GeoLife dataset, demonstrating the robustness, adaptability, and superior performance of MPBS compared to existing baselines.
\end{itemize}

\section{Related Work}

\subsection{Mobile Crowdsensing}

Mobile Crowdsensing (MCS) has emerged as a powerful paradigm for data collection in smart cities, leveraging human participants equipped with mobile devices to perform sensing tasks~\cite{ganti2011mobile, wang2018truthful}. MCS systems typically involve task dissemination, participant recruitment, and data quality evaluation. Numerous studies have explored user recruitment strategies based on incentive mechanisms~\cite{liu2020incentive, di2020novel, xu2019pay}, privacy preservation~\cite{alamri2018privacy, xiao2020privacy, alamri2022privacy}, and participant reputation modeling~\cite{liu2023reputation, xiao2016deadline, zhou2024unknown}.

However, traditional MCS assumes that participants are human users with voluntary task acceptance and unpredictable behavior patterns. These characteristics make it difficult to directly apply MCS methods to autonomous systems like UAVs and UGVs, which exhibit more predictable mobility and controllable availability. In contrast, our work shifts the recruitment paradigm toward autonomous device coordination by treating unmanned devices as intelligent participants in a spatiotemporal task scheduling system~\cite{zhang2022spatio, ma2023utility}.

\subsection{Air-Ground Heterogeneous Collaboration}

In recent years, increasing attention has been paid to heterogeneous collaboration between UAVs and UGVs. These systems aim to leverage the complementary advantages of aerial and ground platforms, such as UAVs’ agility and broad coverage versus UGVs’ stability and endurance~\cite{gupta2015survey}. Applications include joint surveillance, package delivery, and disaster mapping.

Several works have proposed centralized or distributed architectures for air-ground coordination~\cite{zhao2024quadruped, zhao2024ag, rinaldi2024comprehensive, zhang2022spatio}, but often focus on navigation, path planning, or energy efficiency. Relatively few studies emphasize the task recruitment perspective—specifically, how to dynamically select which UAVs or UGVs should participate in a mission based on their spatiotemporal context. Our work fills this gap by providing a predictive recruitment model tailored for heterogeneous unmanned systems~\cite{liu2024optimizing, zhang2022spatio}.

\subsection{Recruitment and Location Prediction}

Recruitment in dynamic environments often depends on accurately estimating the availability and location of participants. For human-centric systems, approaches like trajectory clustering~\cite{yang2018budget}, historical pattern mining~\cite{wang2017efficient}, or probabilistic mobility models~\cite{liu2017prediction} have been applied. In recent work, predictive recruitment for mobile users using Markov models or machine learning has shown promising results in MCS~\cite{zhou2024unknown, xiao2016deadline, wang2018worker}.

For unmanned systems, however, existing work often treats task assignment as a static optimization problem without considering real-time recruitment capability. Moreover, few studies have explored time-varying Markov models to capture periodic mobility trends in autonomous systems. In this paper, we propose a hybrid recruitment framework that combines K-Nearest Neighbor (KNN) classification and a Time-Varying Markov Prediction (TVMP) model, enabling more accurate forecasting of device availability for spatiotemporal task allocation~\cite{wang2022spatiotemporal, wang2017efficient}.

\section{System Model and Problem Formulation}

In this section, we formalize the recruitment problem for unmanned devices by introducing a spatiotemporal task assignment model. The system consists of a set of tasks distributed across different locations and a fleet of autonomous unmanned devices, including UAVs and UGVs, that can be recruited to execute these tasks based on their predicted availability and movement patterns.

\subsection{System Model}

We consider a dynamic task environment composed of the following elements:

\begin{itemize}
  \item \textbf{Tasks}: Let $\mathcal{T} = \{T_1, T_2, ..., T_n\}$ be a set of $n$ tasks. Each task $T_i$ is associated with a location $l_i$, required number of devices $r_i$, and deadline $d_i$.
  
  \item \textbf{Devices}: Let $\mathcal{D} = \{D_1, D_2, ..., D_m\}$ be a set of $m$ unmanned devices. Each device $D_j$ is characterized by its predicted availability window, current location, and behavioral type.
  
  \item \textbf{Base Stations}: Let $\mathcal{B} = \{B_1, B_2, ..., B_k\}$ denote the set of base stations. Each base station $B_b$ maintains recruitment information and coordinates device scheduling within its coverage area.
  
  \item \textbf{Recruitment Factor}: Each base station $B_b$ maintains a recruitment factor $m_b$ based on historical success in recruiting devices, updated over time to reflect recruitment reliability.
\end{itemize}

The system operates in discrete time slots $t \in \{1, 2, ..., T\}$. At each time slot, the goal is to assign available and predicted-to-be-available devices to the appropriate tasks while considering spatiotemporal constraints.

\subsection{Problem Formulation}

The objective is to maximize the total number of successfully recruited devices while satisfying task requirements and respecting system constraints. We define a binary decision variable:

\[
x_{ij} = 
\begin{cases}
1, & \text{if device } D_j \text{ is assigned to task } T_i, \\
0, & \text{otherwise.}
\end{cases}
\]

Our goal is to maximize:

\[
\max \sum_{i=1}^{n} \sum_{j=1}^{m} r_i \cdot x_{ij}
\]

Subject to the following constraints:

\begin{enumerate}
  \item \textbf{Task Assignment Constraint}: Each task can only be assigned up to its required number of devices:
  \[
  \sum_{j=1}^{m} x_{ij} \leq r_i, \quad \forall i \in \{1, ..., n\}
  \]

  \item \textbf{Device Availability Constraint}: Each device can only be assigned to one task at most per time slot:
  \[
  \sum_{i=1}^{n} x_{ij} \leq 1, \quad \forall j \in \{1, ..., m\}
  \]

  \item \textbf{Spatiotemporal Constraint}: A device can only be assigned to a task if it is predicted to be within reach (e.g., within base station coverage) during the task’s execution time:
  \[
  x_{ij} = 0, \quad \text{if } D_j \notin \text{Predicted\_Region}(T_i, d_i)
  \]

  \item \textbf{Base Station Resource Constraint}: The number of devices recruited by base station $B_b$ cannot exceed its operational capacity $c_b$:
  \[
  \sum_{j \in \mathcal{D}_b} \sum_{i=1}^{n} x_{ij} \leq c_b
  \]
\end{enumerate}

This model provides a foundation for the intelligent recruitment and scheduling of unmanned devices. In the next section, we describe how KNN-based classification and time-varying Markov prediction are used to estimate device availability and guide task allocation decisions.

\section{Recruitment Framework and Scheduling Algorithm}

To efficiently schedule unmanned devices for dynamic and distributed tasks, we propose a modular recruitment framework composed of four core components: (A) device behavior classification, (B) spatiotemporal mobility prediction, (C) recruitment expectation estimation, and (D) priority-based task scheduling. Each component is detailed as follows.

\begin{figure}[htbp]
    \centering
    \includegraphics[width=1\linewidth]{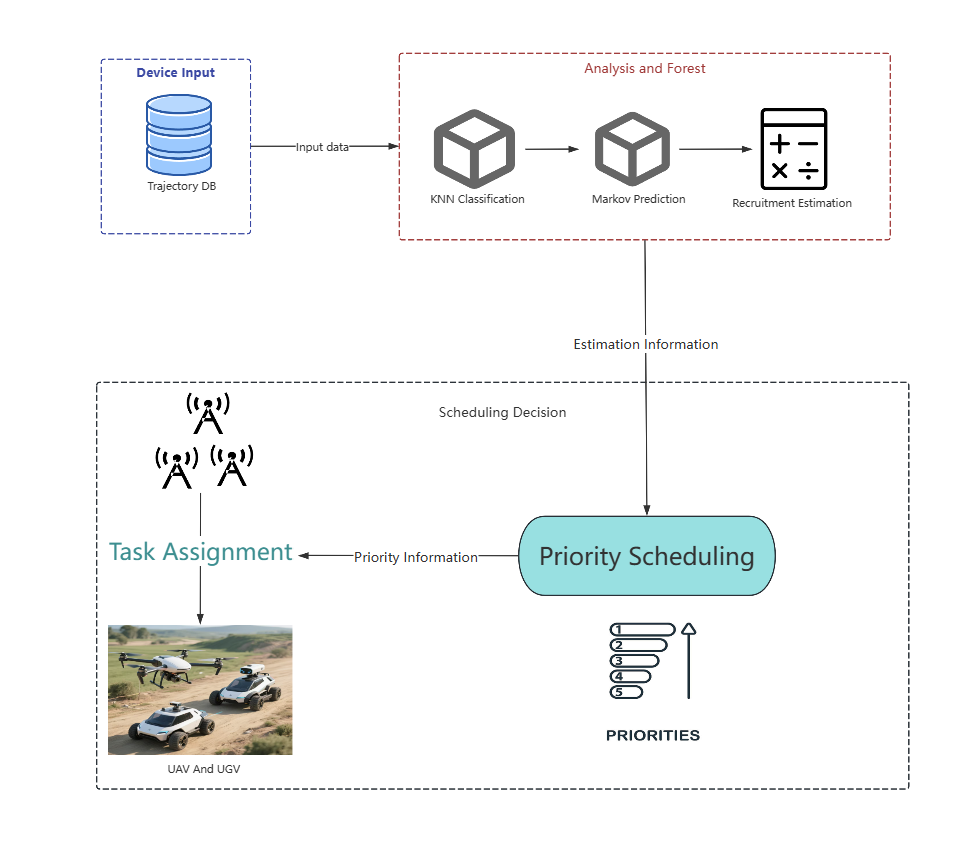}
    \caption{Overall workflow of the proposed MPBS framework.}
    \label{fig:mpbs-framework}
\end{figure}

Figure~\ref{fig:mpbs-framework} illustrates the overall architecture of the proposed MPBS framework. It consists of six key modules: device data input, behavior classification, mobility prediction, recruitment estimation, priority-based scheduling, and task assignment.

\subsection{Device Behavior Classification using KNN}

We classify each unmanned device based on its historical movement behavior using the $k$-Nearest Neighbors (KNN) algorithm. Devices are represented by feature vectors derived from trajectory history, including visit frequency, location entropy, average displacement, and dwell time. They are classified into four types:
\begin{itemize}
    \item \textbf{Regular movers}: predictable and periodic movement patterns.
    \item \textbf{Semi-regular movers}: partially structured but adaptive patterns.
    \item \textbf{Localized movers}: low-range activity with limited transitions.
    \item \textbf{Random movers}: high entropy and no stable trajectory pattern.
\end{itemize}

To evaluate the effectiveness of different classifiers for mobility behavior detection, we compare Decision Tree, Naive Bayes, and K-Nearest Neighbors on our trajectory feature set. As shown in Table~\ref{tab:table1}, KNN achieved the highest classification accuracy of 95.55\%, and is therefore selected as the final model for behavior labeling.

\begin{table}[h]
\caption{Comparison of the effects of various machine learning algorithms\label{tab:table1}}
\centering
\begin{tabular}{|l|l|l|l|}
\hline
\textbf{\begin{tabular}[c]{@{}l@{}}Classification\\ Model\end{tabular}} & 
\textbf{\begin{tabular}[c]{@{}l@{}}Decision\\ Tree\end{tabular}} & 
\textbf{\begin{tabular}[c]{@{}l@{}}Naive\\ Bayesian\end{tabular}} & 
\textbf{\begin{tabular}[c]{@{}l@{}}K-Nearest\\ Neighbors\end{tabular}} \\ 
\hline
Accuracy & 93.75\% & 93.25\% & \textbf{95.55\%} \\
\hline
\end{tabular}
\end{table}

\subsection{Time-Varying Markov Prediction Model}

We propose a behavior-adaptive prediction strategy where different user types are assigned different models to estimate future mobility. Specifically:

\begin{itemize}
  \item For \textit{localized movers}, the preferred next location is determined by selecting the base station with the maximum accumulated dwell time over historical trajectories.
  \item For \textit{semi-regular movers} and \textit{random movers}, second- and third-order Markov models are applied, respectively, to capture trajectory dependencies.
  \item For \textit{regular movers} (e.g., office-like behavioral devices), we use a \textbf{time-varying Markov prediction model}, which considers periodic shifts in mobility trends.
\end{itemize}

To capture these temporal dynamics, we divide a day into four distinct time periods based on observed transition frequency patterns:
\begin{enumerate}
    \item \textbf{0:00–7:00 (Rest Period)}: Low activity, indicative of rest.
    \item \textbf{7:00–10:00 (Commute to Work)}: High-frequency transitions.
    \item \textbf{10:00–17:00 (Work Hours)}: Moderate and stable transitions.
    \item \textbf{17:00–24:00 (Commute Home)}: Second peak in movement.
\end{enumerate}

Each period is assigned an independent transition matrix $P^{(t)}$, where the transition probability is computed as:
\[
  p^{(t)}_{ij} = \frac{n^{(t)}_{ij}}{n^{(t)}_i}
\]
Here, $n^{(t)}_{ij}$ represents the number of transitions from region $i$ to $j$ during time period $t$, and $n^{(t)}_i$ is the total number of transitions originating from region $i$ in $t$.

\begin{figure}[ht]
\centering
\includegraphics[width=1\linewidth]{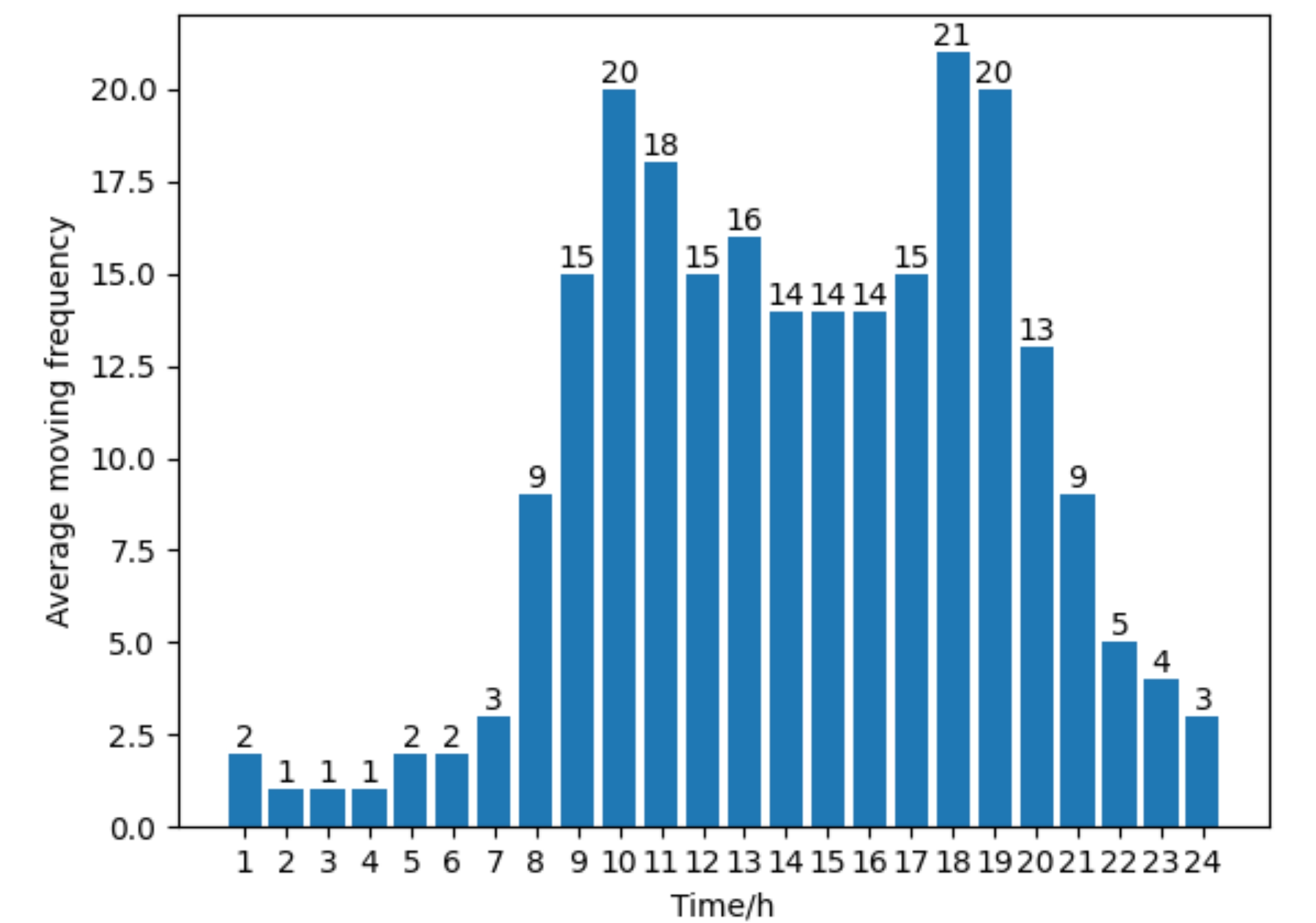}
\caption{Employee position transfer frequency.}
\label{fig_1}
\end{figure}

This segmented modeling approach improves prediction accuracy by aligning with regular daily routines. Fig.~\ref{fig_1} shows the observed bimodal transfer behavior in the raw mobility data of regular movers.

For each time period $t$, we generate a separate Markov matrix and optionally use a second-order extension to capture dependencies across two previous locations. The final output is a probability distribution $\pi^{(t)}_d$ over grid cells for each device $d$ at time $t$.

\subsection{Recruitment Expectation Estimation}

Once each device’s future location distribution is predicted using the Markov model, we further estimate how many devices are expected to be available for a given task at its scheduled execution time. This estimation determines whether a base station has sufficient recruitable resources to handle a task locally or needs to request support.

For a task $T_i$ located at region $l_i$ and scheduled at time $t$, the recruitment expectation $E_i$ is computed as:
\[
E_i = \sum_{j=1}^{m} P^{(t)}_{j,l_i} \cdot \rho_j
\]
where:
\begin{itemize}
    \item $P^{(t)}_{j,l_i}$ is the probability that device $D_j$ will appear in region $l_i$ during time $t$ (estimated via the time-varying Markov model),
    \item $\rho_j$ is the historical reliability score of device $D_j$, representing its likelihood of responding successfully to a task invitation.
\end{itemize}

If $E_i \ge r_i$ (where $r_i$ is the number of devices required by task $T_i$), the task is considered executable with expected available resources. Otherwise, the base station may escalate the request to adjacent stations or reschedule the task.

The estimation process is shown in Algorithm~\ref{alg:recruitment}.

\begin{algorithm}[htbp]
\caption{Recruitment Expectation Estimation}
\label{alg:recruitment}
\KwIn{Predicted location probabilities $P^{(t)}_{dj}$ for all devices;\\
Reliability scores $\rho_d$; Task location $l_i$ and time $t$}
\KwOut{Recruitment expectation $E_i$ for each task $T_i$}

\ForEach{task $T_i$}{
    $E_i \leftarrow 0$\;
    \ForEach{device $D_d$}{
        $p \leftarrow P^{(t)}_{d,l_i}$\;
        $E_i \leftarrow E_i + p \cdot \rho_d$\;
    }
}
\Return{$E_i$ for all tasks}
\end{algorithm}

The algorithm iterates over all $n$ tasks, and for each task, aggregates predicted probabilities from all $m$ devices. Hence, the total time complexity is:
\[
\mathcal{O}(n \times m)
\]
This complexity is acceptable for moderate-scale deployments. In large-scale systems (e.g., thousands of devices), performance can be improved through device pre-filtering, region partitioning, or GPU-parallel implementation.

While this module offers an effective predictive mechanism, it has some limitations:
\begin{itemize}
    \item \textit{Independence Assumption:} The model treats device behaviors independently, which may overlook inter-device interference or collective routing constraints.
    \item \textit{Delayed Reliability Updates:} The $\rho_j$ score updates rely on prior task outcomes, introducing a delay in adapting to short-term device state changes (e.g., battery failure).
    \item \textit{Scalability Challenge:} For very large fleets, the linear complexity can become costly, requiring approximation or pruning strategies.
\end{itemize}

\subsection{Priority-Based Scheduling}

To allocate unmanned devices effectively in scenarios where multiple tasks compete for limited resources, we introduce a priority-based scheduling strategy. This mechanism dynamically ranks tasks based on a composite priority score, which integrates both task urgency and base station reliability.

\textbf{1) Urgency Score Calculation:}

Each task $T_i$ is associated with a deadline $d_i$ and release time $t_i$. We define an urgency score $u_i$ to quantify how close the task is to its deadline:
\[
u_i = \frac{1}{d_i - t_{\text{now}} + \epsilon}
\]
where $t_{\text{now}}$ is the current time slot, and $\epsilon$ is a small positive constant to avoid division by zero. A higher $u_i$ indicates greater urgency.

\textbf{2) Base Station Reliability:}

For each base station $B_j$, we maintain a reliability score $m_j \in [0,1]$ reflecting historical task handling performance. This score is computed from three key metrics:
\begin{itemize}
    \item \textit{Success Rate:} The ratio of successfully completed tasks to attempted ones.
    \item \textit{Average Response Delay:} The average time between recruitment and task execution.
    \item \textit{Utilization Rate:} The frequency at which the station actively dispatches tasks.
\end{itemize}
These metrics are normalized and combined to derive $m_j$, which reflects the base station’s efficiency and dependability.

\textbf{3) Composite Priority Score:}

We define the final priority score $P_i$ for each task $T_i$ (assigned to base station $B_j$) as:
\[
P_i = \alpha \cdot u_i + \beta \cdot m_j
\]
where $\alpha$ and $\beta$ are tunable weights, and $\alpha + \beta = 1$. This formulation allows balancing task urgency and infrastructure performance. In practice, higher $\alpha$ (e.g., 0.8) favors time-sensitive tasks, while higher $\beta$ (e.g., 0.5) increases trust in historically reliable base stations.

\textbf{4) Scheduling and Device Allocation:}

All pending tasks are sorted in descending order of $P_i$. Available devices are allocated to the highest-priority tasks, subject to predicted availability and capacity constraints. The allocation respects:
\begin{itemize}
    \item The number of devices required by each task.
    \item Each device’s predicted availability (from the Markov model).
    \item Each device can only be assigned to one task per time slot.
\end{itemize}

This mechanism ensures that:
\begin{itemize}
    \item Urgent tasks are handled in a timely manner.
    \item Base stations with poor performance are penalized in scheduling priority.
    \item The overall task success rate and resource efficiency are improved.
\end{itemize}

The priority scores and reliability factors are updated periodically to reflect real-time changes in task patterns and execution results, enabling the system to adaptively optimize scheduling decisions. The algorithm is presented below.

\begin{algorithm}[htbp]
\caption{Priority-Based Task Scheduling}
\KwIn{Urgency scores $u_i$, base station reliability scores $m_j$;\\
Task list $\mathcal{T} = \{T_1, ..., T_n\}$;\\
Weights $\alpha$, $\beta$}
\KwOut{Sorted priority queue $\mathcal{T}'$ and device allocation plan}

\ForEach{task $T_i$ from base station $B_j$}{
    Compute priority score: $P_i = \alpha \cdot u_i + \beta \cdot m_j$\;
}
Sort all tasks $\mathcal{T}$ in descending order of $P_i$ to get $\mathcal{T}'$\;

\ForEach{task $T_i$ in $\mathcal{T}'$}{
    \ForEach{available device $D_d$ predicted to appear at $l_i$}{
        \If{$D_d$ not assigned to other tasks \textbf{and} $r_i$ not reached}{
            Assign $D_d$ to $T_i$\;
            Update counters and availability\;
        }
    }
}
\Return{Task-to-device allocation result}
\end{algorithm}

\section{Experiments and Evaluation}

To evaluate the effectiveness and efficiency of the proposed MPBS (Mobility-aware Prediction and Behavior-based Scheduling) framework, we conduct a comprehensive simulation study based on a real-world GPS mobility dataset. This section introduces the dataset used, simulation setup, comparison algorithms, and evaluation criteria.

\subsection{Dataset and Simulation Configuration}

We use the publicly available GeoLife GPS Trajectory Dataset~\cite{zheng2010geolife}, which contains 17,621 trajectories collected by 182 users over a period of five years, primarily in Beijing, China. Each trajectory includes timestamped latitude and longitude coordinates with high temporal resolution (1–5 seconds), enabling fine-grained mobility modeling.

To simulate unmanned devices in a 2D urban environment, we discretize the city space into a $10 \times 10$ region grid. All trajectory data are mapped into grid regions, and segmented into time slots of 15 minutes. For each time slot $t$, we construct a transition matrix $P^{(t)}$ using observed region transitions, which supports time-varying Markov prediction.

We generate synthetic tasks at random grid locations with variable requirements, including release time $t_i$, deadline $d_i$, number of required devices $r_i$, and geographic position $l_i$. Each simulated UAV/UGV device follows location transitions derived from the GeoLife mobility patterns, with pre-assigned behavior types and associated reliability scores $\rho_j$.

We simulate 300 devices and 1000 tasks per run. Each simulation spans 96 time slots (one day), and is repeated 10 times with different seeds to ensure statistical robustness.

\begin{table}[h]
\caption{GeoLife Trajectory Dataset Statistics\label{tab:table-dataset}}
\centering
\begin{tabular}{|l|l|}
\hline
\textbf{Attribute}              & \textbf{Value}              \\ \hline
Number of users                 & 182                         \\ \hline
Total trajectories              & 17,621                      \\ \hline
Average sampling interval       & 1--5 seconds                \\ \hline
Total duration span             & Apr 2007 -- Aug 2012        \\ \hline
Total GPS points                & Over 24 million             \\ \hline
Primary region                  & Beijing, China              \\ \hline
\end{tabular}
\end{table}

\subsection{Comparison Algorithms and Evaluation Metrics}

To evaluate the performance of our framework, we compare the MPBS algorithm against three representative baselines:

\begin{itemize}
    \item \textbf{Greedy}~\cite{zhu2018spatiotemporal}: A classic utility-based allocation strategy. It selects devices one by one that yield the highest immediate marginal gain in coverage until the task's requirement is fulfilled. While computationally efficient, it neglects device availability in future time slots and lacks long-term planning.

    \item \textbf{HSF (Heuristic Score Filtering)}~\cite{zhu2018spatiotemporal}: This heuristic algorithm assigns a score to each device based on its coverage-to-cost ratio. Devices are sorted and activated in descending order, and the task coverage is re-evaluated after each assignment. Devices offering little marginal benefit are filtered out. HSF balances efficiency and sensing cost, but remains myopic in dynamic environments.

    \item \textbf{PPO (Proximal Policy Optimization)}~\cite{schulman2017proximal}: A reinforcement learning-based baseline that models the task allocation as a sequential decision-making process. PPO updates a stochastic policy using clipped surrogate objective functions and performs well in high-dimensional action spaces. It learns task-device mappings from experience but requires extensive training data and environment interaction.

    \item \textbf{MPBS (Mobility-aware Prediction and Behavior-based Scheduling)}: Our proposed algorithm combines multiple predictive and adaptive modules to enable proactive task-device matching. It consists of:
    \begin{itemize}
        \item \textit{Behavior Classification}: Each device is classified into one of four behavioral types (e.g., regular, random) via a KNN model trained on trajectory features.
        \item \textit{Time-Varying Markov Prediction}: A tailored prediction strategy is applied based on behavior type. Regular devices use segmented time-slot Markov matrices to anticipate mobility more accurately.
        \item \textit{Expectation Estimation}: Using predicted location probabilities and historical reliability scores, MPBS estimates the expected number of recruitable devices per region and task.
        \item \textit{Priority-Based Scheduling}: A composite score ranks tasks based on urgency and base station reliability. Tasks with higher scores receive priority in device allocation.
    \end{itemize}

\end{itemize}

We evaluate all algorithms using the following standard metrics:

\begin{itemize}
    \item \textbf{Task Completion Rate (TCR)}: The percentage of tasks that are completed before their deadlines.
    \item \textbf{Average Response Time (ART)}: The mean time delay from task generation to successful device assignment.
    \item \textbf{Device Utilization (DU)}: The fraction of devices actively involved in executing tasks.
\end{itemize}

Each experiment is repeated multiple times with different random seeds, and mean performance with confidence intervals is reported.

\subsection{Analysis of Simulation Results}

In our evaluation, we analyzed the effects of both node scale and device heterogeneity on the performance of MPBS, alongside three baseline algorithms (Greedy, HSF, and PPO). The goal was to examine when maintaining a fixed ratio among UAVs, UGVs, and base stations, how varying the number of participating unmanned devices would impact task completion, response efficiency, and overall resource utilization.

\begin{figure}[htbp]
\centering
\includegraphics[width=0.48\textwidth]{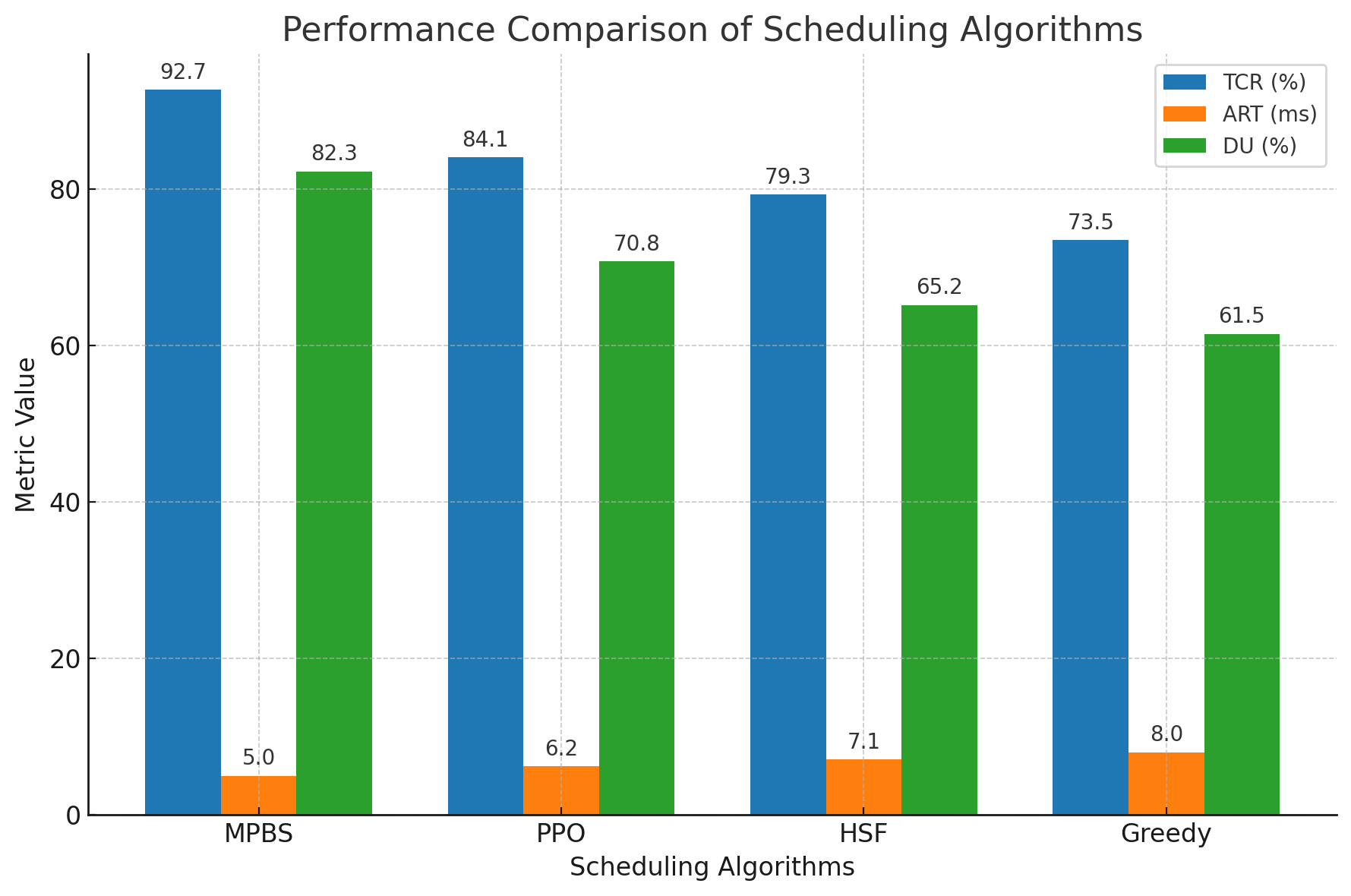}
\caption{Performance Comparison of Scheduling Algorithms: MPBS achieves superior TCR, lower ART, and higher DU compared to baseline methods.}
\label{fig:performance-comparison} 
\end{figure}

This question is critical for designing scalable systems, as the number of active devices not only determines the task execution potential but also influences scheduling efficiency and communication overhead. To simulate realistic conditions, we maintained a fixed ratio of UAV:UGV:Base Station = 5:5:1 and varied the total node count from 20 to 80. The results, presented in Figure~\ref{fig:performance-comparison}, demonstrate that MPBS consistently outperforms the baselines across all node scales, particularly under constrained-resource scenarios, where traditional algorithms struggle with effective task coverage and timely response.

To further examine the adaptability of MPBS in resource-constrained environments, we compare its base-station-centric scheduling strategy with two traditional task-based prioritization methods: Earliest Deadline First (EDF) and Least Slack First (LSF). Unlike EDF and LSF, which rank tasks purely based on timing, our method considers the historical performance of base stations—such as recruitment success rate, task completion rate, and average execution delay—to assign priorities.

We model the assignment as an optimization problem with constraints on device availability and station capacity. Let $T_i$ be the priority of task $i$, $R_i$ the number of required devices, $S_j$ the capacity of base station $j$, and $X_{ij}$ a Boolean variable indicating if task $i$ is assigned to base station $j$.

Objective:
\[
\text{Maximize } \sum_{i=1}^n \sum_{j=1}^m R_i \cdot X_{ij}
\]
Subject to:
\[
\sum_{j=1}^{m} X_{ij} = 1, \quad \forall i
\quad \text{and} \quad
\sum_{i=1}^n R_i \cdot X_{ij} \leq S_j, \quad \forall j
\]

We further define evaluation metrics as:
\begin{itemize}
  \item Number of successful recruits: $NP = \sum_{i=1}^n \sum_{j=1}^m R_i \cdot X_{ij}$
  \item Task completion rate: $CR = \frac{\text{Completed Tasks}}{\text{Total Tasks}}$
  \item Average task time: $AT = \frac{\sum_{i=1}^n t_i}{n}$
\end{itemize}

\begin{figure}[h]
    \centering
    \includegraphics[height=0.3\textheight,width=0.999\linewidth]{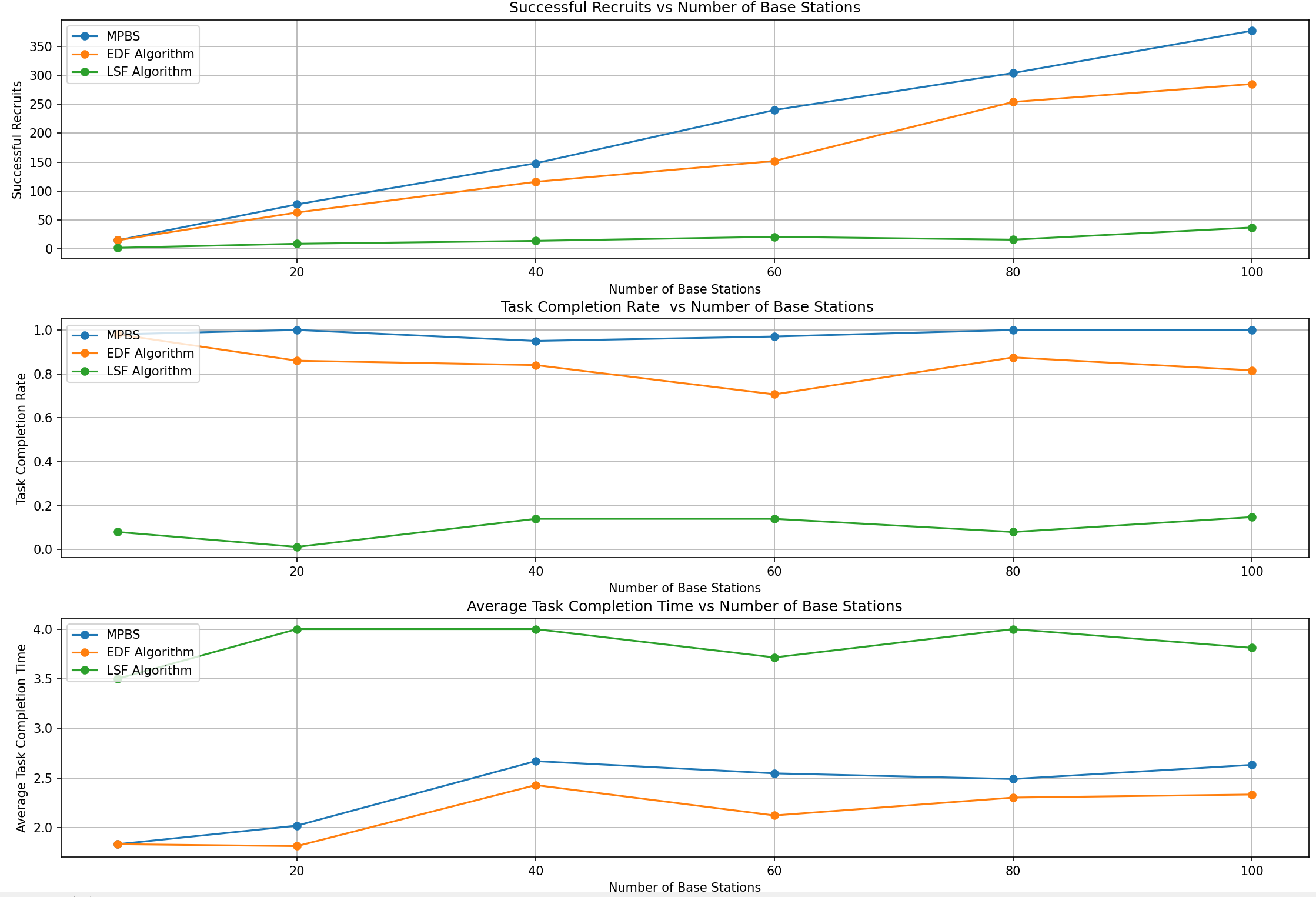}
    \caption{Comparison of base-station-centric MPBS scheduling vs EDF and LSF strategies under varying base station/task ratios.}
    \label{fig:enter-label}
\end{figure}

Figure~\ref{fig:enter-label} compares the average recruitment success, task completion rate, and completion time of our base station-driven MPBS scheduling model with EDF and LSF under various base station/task ratios.

The results show that MPBS achieves comparable or better average task completion times than EDF while outperforming both EDF and LSF significantly in terms of completion rate. Especially when the number of base stations is much smaller than the number of tasks, LSF tends to fail due to suboptimal resource allocation. In contrast, MPBS maintains a task completion rate above 97\%, demonstrating superior efficiency and robustness in dynamic, resource-limited conditions.
consistently maintains high performance, demonstrating scalability and robustness.

\section{Conclusion and Future Work}

In this paper, we proposed MPBS, a task recruitment and scheduling framework for unmanned devices. MPBS effectively addresses two core challenges:
(1) accurate location prediction for mobile devices through behavior-aware classification and time-varying Markov models, and
(2) efficient task allocation under constrained resources using priority-based scheduling and recruitment estimation. In future work, we will focus on more realistic deployment factors, such as communication latency, energy constraints, and fault tolerance. Moreover, we plan to implement a real-world prototype to validate the performance of MPBS in practical UAV/UGV task orchestration scenarios.

\bibliographystyle{IEEEtran} 
\bibliography{ref}  

\end{document}